# How resampling method can help us: Improving the Delta model

Jacques Savoy
University of Neuchatel
2000 Neuchatel, Switzerland
Jacques.Savoy@unine.ch

**Abstract**

Delta has been successfully employed as an authorship attribution method in various contexts, although its effectiveness is not always perfect. Since it is based on the most frequent words, its results can be readily interpreted and explained in plain English. In this study, we propose improving its performance by applying a resampling method to the selected stylistic features. The results show that this approach increases the accuracy of the model while preserving the interpretability of its outcomes. However, the main drawback of the proposed technique is the additional computational time required to obtain more accurate results.

## 1 Introduction

In various authorship attribution studies (Savoy, 2020; Kreuz, 2023), emphasis is placed on a mean performance measure, usually computed by averaging the results obtained across a test sample (Zangerle et al., 2026). In other cases, precision and recall are reported separately, but an overall average measure is ultimately provided. Such approaches are appropriate when multiple decisions must be made and average performance provides a meaningful assessment of a model's behaviour. Nowadays, numerous models have been proposed for identifying the author of a text, together with a wide range of approaches to representing stylistic features (e.g., words, lemmas, word or character sequences, hapax legomena density, sentence length, sequences of part-of-speech tags, spelling, and layout).

However, in many real-world situations, only a single decision must be reached, for example, when examining a newly disputed novel (Tuzzi & Cortelazzo, 2018; Savoy, 2018, 2026), a poem (Thisted & Efron, 1987), a suicide note, or a last will and testament (Olsson, 2018; Kreuz, 2023). For such applications, particularly in a forensic context, the proposed method must be considered reliable and must have been tested under a variety of conditions, with its average performance clearly established (Chaski, 2013). Moreover, both the underlying computational procedures and the reasoning supporting the final attribution must be explained in plain English.

Although these requirements may seem self-evident, the history of authorship attribution includes several notable failures. For example, the CUSUM model (Morton, 1978), based on the cumulative sums of stylistic measurements, was initially recommended as an effective method for authorship identification. This technique was even adopted as forensic evidence in several English courts. However, its effectiveness was soon questioned, and its accuracy was found to be unacceptable (Hilton & Holmes, 1993; Hardcastle, 1997).

Another well-known case concerns the poem *A Funeral Elegy*, signed with the initials W.S. The identification of its author proved particularly controversial. Based on rare-word analysis and subsequent tests, the poem was initially attributed to William Shakespeare (Foster, 1997), before its authorship was eventually assigned to John Ford (Grieve, 2007).

When a single but consequential authorship attribution decision must be made, the main objective of this study is to propose and evaluate a resampling-based technique for improving the accuracy of an authorship attribution model. To demonstrate the advantages and limitations of the proposed technique, we have selected the Delta method (Burrows, 2002). This approach continues to be successfully applied in various contexts, which justifies further research aimed at improving its effectiveness. Nevertheless, its overall performance is not flawless, and some of its parameters remain open to discussion (e.g., the number of stylistic features to be considered).

The remainder of this paper is organized as follows. The next section describes the Danish corpus used in our investigations. Section 3 presents the Delta model, which serves as the main authorship attribution method examined in this study. Section 4 introduces and evaluates the proposed resampling approach, which is applied in conjunction with Delta to improve its effectiveness. Finally, the conclusion summarizes the main findings of this study.

## 2 Corpus of Danish Novels

To provide an experimental basis for this study, a Danish literary collection was compiled by Danish colleague. Table 1 lists the 14 Danish writers included in the corpus, together with their names, genres, and the number of novels available for each author. The values reported in the last three columns are calculated from the concatenation of all the books written by each author.

In the column labelled "Length (in tokens)," the total is computed by treating as tokens all words[1], punctuation symbols, numbers, and alphanumeric terms (e.g., IBM360). In the next column, only words, defined as sequences of letters, are counted, regardless of whether they appear in uppercase or lowercase. The final column reports the vocabulary size, that is, the total number of distinct word-types used by each author.

Several important considerations should be taken into account. First, the corpus contains approximately 2,106,150 tokens, a total roughly nine times greater than the length of the longest English novel, Melville's *Moby-Dick*. Second, all the novels were published within a relatively short period of approximately ten years. Third, the text genres represented in the corpus are broadly comparable, as the books consist mainly of novels intended for adult readers (some of the subgenres represented in individual books are listed in Table A.1). Fourth, each novel contains more than 10,000 tokens, which is considered sufficient to achieve a relatively high degree of reliability with the selected authorship attribution models.

[1] Since the term *word* is potentially ambiguous, some clarification is necessary. In the sentence "I saw a man with a saw," there are seven word-tokens (or simply tokens), but only five distinct word-types (or types), namely {I, saw, a, man, with}.

Before analyzing the corpus, some preprocessing is required. First, all uppercase letters are converted to lowercase. For example, *Og* becomes *og* (and). However, our tokenizer identifies words written entirely in uppercase letters (e.g., DENMARK, WHO), which are left unchanged. The same procedure is applied to word-types that consistently occur with an initial capital letter, such as Matthew, Marie, and Elliot. As these examples illustrate, such word types are generally proper names.

**Table 1.** Descriptive statistics for the novelists representing in the corpus

| Author name | First name | Genre | Novel | Length (in tokens) | Length (in words) | Vocabulary |
|---|---|---|---|---|---|---|
| Blædel | Sara | F | 2 | 213,326 | 181,738 | 181738 |
| Conradsen | Keld | M | 2 | 244,840 | 210,427 | 16,598 |
| Gazan | Sissel-Jo | F | 2 | 214,291 | 174,354 | 13,841 |
| Gerhardt | Maria | F | 2 | 56,627 | 47,011 | 7,567 |
| Guldager | Katrine | F | 2 | 194,482 | 164,053 | 14,301 |
| Halfdan | Finn | M | 2 | 281,606 | 226,622 | 16,769 |
| Harvig | Steen | M | 2 | 186,570 | 151,245 | 13,128 |
| Høeg | Tine | F | 2 | 40,779 | 38,751 | 5,973 |
| Jessen | Ida | F | 2 | 87,016 | 72,563 | 8,331 |
| Korsgaard | Thomas | M | 2 | 165,634 | 131,043 | 9,403 |
| Nordbo | Mads Peder | M | 3 | 325,130 | 261,289 | 12,810 |
| Petersen | Eddie | M | 2 | 183,674 | 158,065 | 14,393 |
| Sabroe | Morten | M | 2 | 222,426 | 177,688 | 13,376 |
| Tellerup | Malte | M | 2 | 99,079 | 81,358 | 9,712 |

As Danish is less frequently used in natural language processing than some other languages, a brief overview of its main linguistic characteristics may be useful. Danish belongs to the North Germanic branch of the Germanic language family, together with Norwegian and Swedish. It has a relatively rich inflectional morphology (e.g., *bil/biler* (car/cars)). Its basic word order follows the subject–verb–object (SVO) pattern, with the finite verb usually occupying the second position in main clauses. Danish has two grammatical genders, common and neuter, and two numbers, singular and plural. The plural is often marked by the suffixes *-er* or *-e*. A distinctive feature of Danish is that definiteness can be expressed by a suffix attached to the noun, in both the singular and the plural, as illustrated by *bilen* (the car) and *bilerne* (the cars).

As in all natural languages, exceptions occur, for example, in noun inflection (*fod/fødder* (foot/feet)) and verb conjugation (*tager/tog* (takes/took)). Lexical ambiguities can also arise. For instance, *I* can function either as a pronoun (the second-person plural form) or as a preposition[2]. Similarly, *det* can function either as a determiner or as a pronoun. Three additional characters are particularly characteristic of Danish orthography: *ø* (historically related to oe), *å*

[2] The tokenizer detects *I* (pronoun) and *i* (preposition) and thus transforms all *I* (pronoun) into lowercase letter (as the pattern 'The' and 'the' that occurs in English).

(historically related to aa), and *æ*.

Some general lexical statistics provide an overview of the lexical characteristics of the corpus. The 20 most frequent word-types in Danish are {*og, det, i, at, han, på, jeg, var, en, er, ikke, til, hun, der, med, så, den, af, for, havde*}. As can be observed, these items generally do not convey specific lexical or topical meaning. Rather, they belong to functional categories, including determiners (*det, en*), prepositions (*af, i*), conjunctions (*og, eller*), pronouns (*jeg, dig, han, hun*[3]), and auxiliary or highly frequent verb forms (*var*, *havde*). The linguistic fingerprint of an author is often strongly associated with such closed-class words, because their use is largely unconscious and is less directly related to the topic or content of a document.

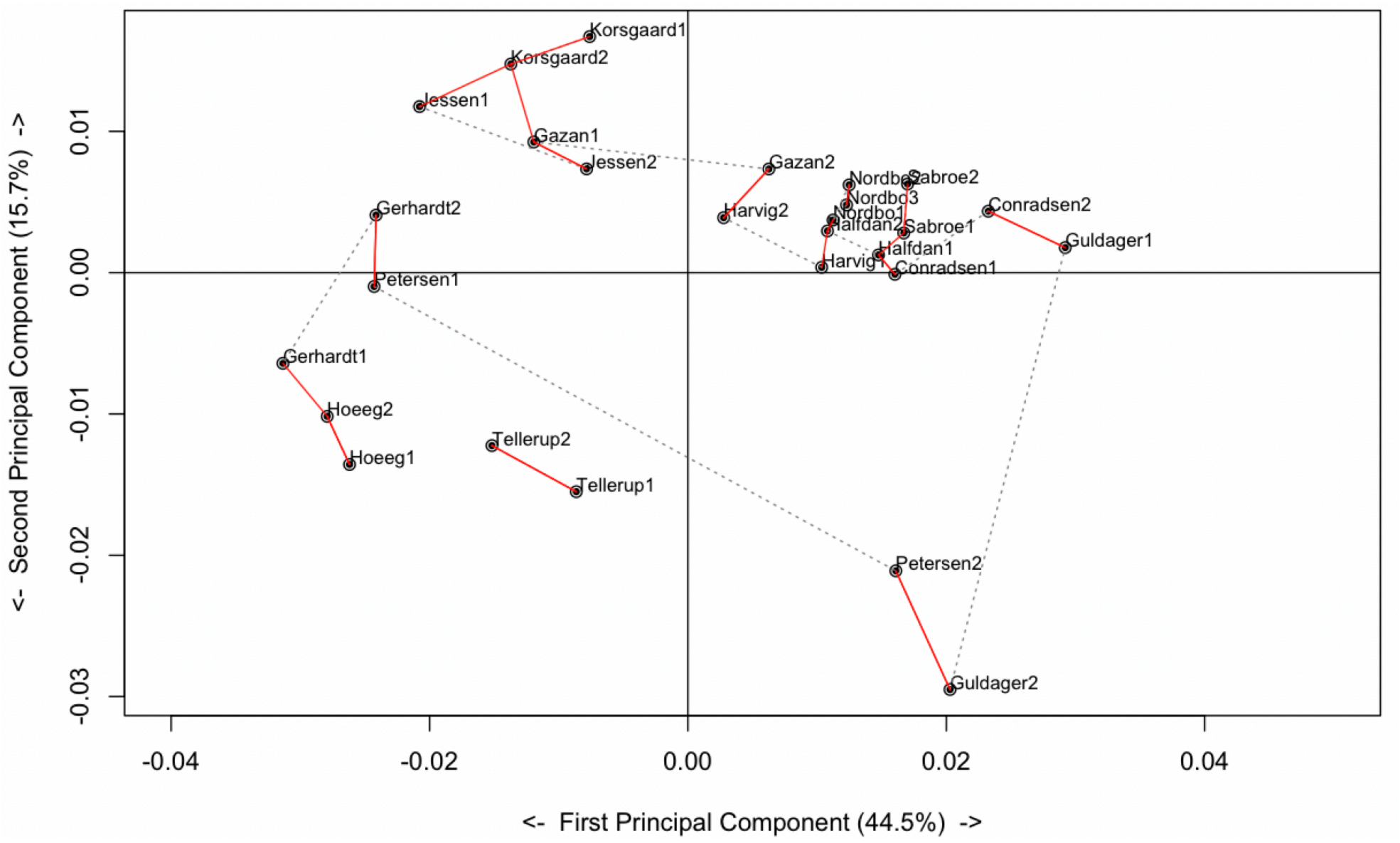

**Figure 1.** PCA projection of Danish novelists based on the 200 most frequent words (MFW)

Finally, to obtain an initial overview of the stylistic differences among the Danish novelists, Figure 1 presents a Principal Component Analysis (PCA) projection based on the 200 most frequent words (MFW) (Binongo & Smith, 1999; Craig & Kinney, 2009; Tuzzi & Cortelazzo, 2018). The projection is based on the first two principal components, which account for 44.5% and 15.7% of the total variation among the novelists, respectively. Thus, the two-dimensional representation captures 60.2% of the overall variation, while the remaining 39.8% is not represented in the figure.

In this figure, each novel is connected to its nearest neighbor by a solid line. This connection is determined solely on the basis of the first two principal components displayed in Figure 1. In addition, a gray dotted line connects each novel to the second book written by the same author. Figure 1 shows that this simple nearest-neighbor attribution scheme produces several errors. For example, the two novels by Guldager and those by Petersen are separated by relatively large distances. For several other authors, the nearest-neighbor assignment is also incorrect (e.g., Gerhardt, Jessen, Gazan, and Harvig).

[3] As already observed in other corpora, the masculine form (*han*) occurs more frequently than the feminine form (*hun*).

It is important to note, however, that this representation does not preserve orthonormal distance scaling, meaning that the units of distance are not equivalent along the two axes. In the present case, the horizontal axis spans a greater range of variation than the vertical axis. Consequently, visual distances between points should be interpreted with caution.

## 3 Authorship Attribution Based on Delta

To identify the true author of a document, several authorship attribution models have been proposed. Among those, Burrows' Delta has been found effective in different studies. Originally proposed by Burrows (2002), several variants have subsequently been proposed and evaluated (Hoover, 2004a, 2004b; Burrows, 2007; Hoover, 2007), while its theoretical properties have also been investigated (Argamon, 2008). The model has been employed in numerous authorship attribution studies (Jockers et al., 2008; Savoy, 2012, 2016; Eder, 2015; Evert et al., 2017; Kocher & Savoy, 2018) and even to evaluate machine translation systems (Mikros et al., 2026). Moreover, Jockers and Witten (2010) showed that the Delta method can outperform Support Vector Machine (SVM) approaches in authorship attribution.

This attribution model is based on a subset of the vocabulary and considers only the most frequent word-types (MFW), which consist mainly of function words. The number of terms to be included (e.g., word-types or short sequences of letters) is not precisely defined; however, values between 200 and 500 terms are commonly used. These terms are selected without making any explicit reference to the query document.

More precisely, for each selected term $t_i$ in each text $T_j$, its relative term frequency $rtf_{ij}$ is computed. This value is then standardized using the mean (denoted $\overline{rtf_i}$) and standard deviation ($s_i$) of the corresponding term across all texts in the reference corpus. The resulting standardized value, denoted Z score$(t_{ij})$, is defined according to Equation 1.

$$\text{Z score}\left(t_{ij}\right) = \left(\text{rtf}_{ij} - \overline{rtf_i}\right) / s_i \quad (1)$$

Given a query text $Q$, an a text $A_k$ written by author A, and a set of words $t_i$, for $i = 1, 2, \ldots, m$, the Delta distance between $Q$ and $A_k$, denoted $D(Q, A_k)$, is calculated according to Equation 2. This formulation corresponds to the Manhattan distance with, in addition, the division by $m$.

$$\Delta(Q, A_k) = 1/m \cdot \sum_{i=1}^{m} \left| \, Z\, score\left(t_{iQ}\right) - Zscore\left(t_{ik}\right) \right| \quad (2)$$

When, for a given term, the two $Z$ scores are large and have opposite signs, the distance increases. In this situation, one author tends to employ the corresponding word more frequently than the corpus mean, whereas the other tends to use it less frequently. Conversely, when the $Z$ score values are very similar for all terms, the distance between the two texts is reduced. The author of text $A_k$ yielding the smallest distance, among all possible writers is therefore identified as the most likely author of the query text.

Moreover, differences in Z scores can be used to provide a plain English interpretation of stylistic distance. When two terms exhibit only a small difference in their Z scores, they occur with similar relative frequencies in both texts, indicating a lexical similarity. Conversely, a large difference suggests a stylistic disparity between the two documents.

To apply the Delta model to the present corpus, the set of stylistic features must first be defined. This set consists of the *m* most frequent word-types (MFW), identified by considering all the books included in the corpus

Finally, the Manhattan distance measure is selected to quantify the stylistic distance between two texts, given its good empirical performance in previous studies. The resulting distance ranges from zero (e.g., the distance between a text and itself) to positive values that may exceed 1.0. As the number of stylistic features increases, the Manhattan distance will also increase, simply because more dimensions are taken into account (see Equation 2). Consequently, the absolute magnitude of a distance value cannot be interpreted in isolation. For example, a distance of 0.75 may be considered either large or small, depending on the other distance values obtained. The appropriate approach is therefore to interpret each distance value in relation to the other values within the same ranking.

As a first experiment, all novels were selected in turn, using a round-robin procedure, as query documents. The Delta model was then applied to the remaining books, taking advantage of the fact that each author is represented by at least two texts. Thus, for each query novel, the expected result is that the second book by the same author should appear in the first position in the ranking. Table 2 reports the three highest-ranked results for the 29 books included in the Danish corpus. For 27 of the 29 novels (or 93.1%[4]), Delta produces the expected result. However, for two books (highlighted in bold in Table 2), another writer is ranked first. The first misclassification concerns Guldager1, for which Gazan2 is identified as the closest stylistic match. The second concerns Høeg1, which is matched with Tellerup2.

Table 2 also illustrates the stylistic consistency that can be observed across different novels by the same author. For example, Nordbo is represented by three books, and in all three cases, the other two novels appear in the first two positions of the ranking, with relatively small Delta distances.

A closer examination of the distance values reported in Table 2 indicates that, when an incorrect attribution is obtained, the difference between the distances associated with the first and second-ranked candidates is relatively small compared with the corresponding difference in cases of correct attribution. For example, for Guldager1, the difference is 0.789 − 0.774 = 0.015 (or 1.9%). By contrast, for Guldager2 (correct attribution), the difference is 1.043 − 0.803 = 0.240 (or 23%). On average, the relative difference between the distances associated with the two highest-ranked candidates is 23.7% for correct attributions, compared with only 1% for incorrect ones. This observation may provide a useful indication when dealing with challenging authorship attribution problems.

---

[4] For a fair assessment, the estimated accuracy should be adjusted according to the add-one principle (Manning & Schütze, 1999). In the present case, the estimated accuracy is therefore not simply 27/29 = 93.1%, but rather (27 + 1)/(29 + 2) = 90.3%.

**Table 2.** Author ranking obtained using Delta (200 MFW)

| \ Rank | 1 | | 2 | | 3 | |
|---|---|---|---|---|---|---|
| Novel | Novel | Distance | Novel | Distance | Novel | Distance |
| Blædel1 | Blædel2 | 0.516 | Gazan2 | 0.819 | Nordbo3 | 0.867 |
| Blædel2 | Blædel1 | 0.519 | Gazan2 | 0.798 | Nordbo3 | 0.898 |
| Conradsen1 | Conradsen2 | 0.552 | Guldager1 | 0.780 | Gazan2 | 0.878 |
| Conradsen2 | Conradsen1 | 0.557 | Guldager1 | 0.813 | Jessen2 | 0.909 |
| Gazan1 | Gazan2 | 0.630 | Jessen2 | 0.808 | Jessen1 | 0.835 |
| Gazan2 | Gazan1 | 0.619 | Halfdan1 | 0.703 | Jessen2 | 0.747 |
| Gerhardt1 | Gerhardt2 | 0.927 | Jessen1 | 0.963 | Petersen1 | 0.999 |
| Gerhardt2 | Gerhardt1 | 0.945 | Jessen1 | 0.988 | Jessen2 | 1.011 |
| **Guldager1** | **Gazan2** | 0.774 | Conradsen1 | 0.789 | *Guldager2* | 0.793 |
| Guldager2 | Guldager1 | 0.803 | Gerhardt1 | 1.043 | Høeg1 | 1.053 |
| Halfdan1 | Halfdan2 | 0.410 | Gazan2 | 0.703 | Nordbo3 | 0.757 |
| Halfdan2 | Halfdan1 | 0.410 | Harvig1 | 0.733 | Nordbo3 | 0.772 |
| Harvig1 | Harvig2 | 0.431 | Halfdan2 | 0.751 | Halfdan1 | 0.783 |
| Harvig2 | Harvig1 | 0.430 | Halfdan2 | 0.794 | Halfdan1 | 0.824 |
| **Høeg1** | **Tellerup2** | 0.985 | Petersen1 | 0.989 | *Høeg2* | 1.036 |
| Høeg2 | Høeg1 | 1.114 | Tellerup2 | 1.416 | Petersen1 | 1.417 |
| Jessen1 | Jessen2 | 0.592 | Gazan2 | 0.752 | Gazan1 | 0.825 |
| Jessen2 | Jessen1 | 0.596 | Gazan2 | 0.754 | Gazan1 | 0.801 |
| Korsgaard1 | Korsgaard2 | 0.454 | Gazan2 | 0.840 | Jessen1 | 0.887 |
| Korsgaard2 | Korsgaard1 | 0.458 | Gazan2 | 0.923 | Jessen1 | 0.956 |
| Nordbo1 | Nordbo2 | 0.364 | Nordbo3 | 0.451 | Halfdan1 | 0.809 |
| Nordbo2 | Nordbo1 | 0.364 | Nordbo3 | 0.411 | Halfdan1 | 0.806 |
| Nordbo3 | Nordbo2 | 0.406 | Nordbo1 | 0.446 | Halfdan1 | 0.758 |
| Petersen1 | Petersen2 | 0.703 | Jessen1 | 0.869 | Høeg1 | 0.959 |
| Petersen2 | Petersen1 | 0.710 | Tellerup2 | 1.034 | Høeg1 | 1.062 |
| Sabroe1 | Sabroe2 | 0.411 | Halfdan2 | 0.793 | Nordbo3 | 0.843 |
| Sabroe2 | Sabroe1 | 0.416 | Halfdan2 | 0.839 | Halfdan1 | 0.894 |
| Tellerup1 | Tellerup2 | 0.739 | Høeg1 | 1.087 | Petersen2 | 1.118 |
| Tellerup2 | Tellerup1 | 0.732 | Høeg1 | 0.984 | Petersen2 | 1.047 |

When the number of MFWs is varied (specifically, to 100, 250, 300, 500, or 800), the overall performance remains similar to that reported in Table 2. More precisely, with 250 or 500 MFWs, Høeg1 is incorrectly matched with Tellerup2, while Guldager1 is also incorrectly assigned when using 100 or 800 MFWs. In addition, Gerhardt1 is matched with Tellerup2 when using 500 or 800 MFWs. These six inconsistencies across the different parameter settings can be explained, at least in part, by examining Figure 1, where these novels appear relatively distant from the second novel written by the same author.

In a second experiment, each novel is divided into a series of fixed-length, non-overlapping segments, for example, chunks of 5,000 tokens (words, numbers, and punctuation symbols). In a round-robin fashion, each segment is compared with the author profiles of the 14 novelists. This identification procedure is known as Rolling Delta (Eder, 2016). In the present study, Guldager1 was divided into 22 non-overlapping segments of 5,000 tokens each. This novel was selected because of the difficulty in identifying the correct second book by the same author, as reported in Table 2. Similarly, Guldager2 was divided into 16 segments. For each segment, the Delta model was applied using the 200 most frequent word-types (MFW) as stylistic features.

The results of this Rolling Delta procedure are presented in Figure 2. The segments from Guldager1 are represented by circles, while those from Guldager2 are represented by triangles. The vertical axis indicates the Delta distance calculated for each segment. The mean Delta distance for each novel is indicated by a solid horizontal line (1.40 for Guldager1 and 1.37 for Guldager2). In addition, whenever Guldager is not ranked first for a given segment, the corresponding point is highlighted with a square.

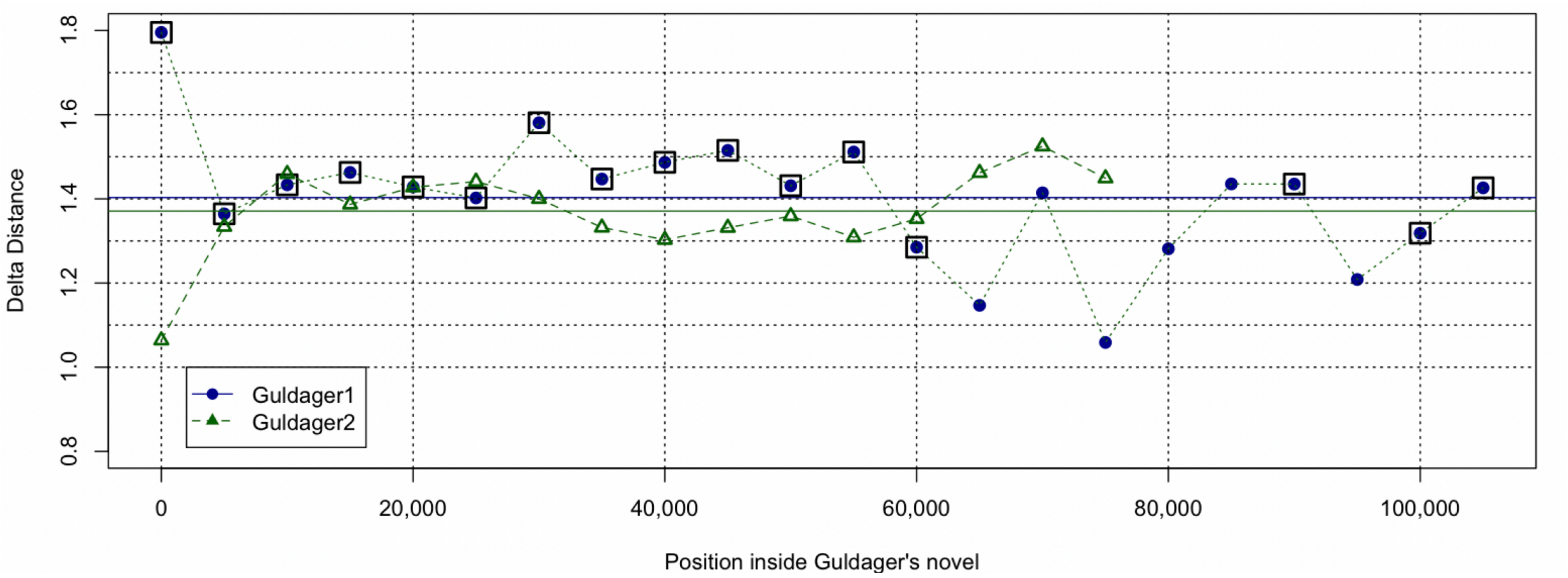

**Figure 2.** Rolling Delta distance values (200 MFW) for Guldager's novels

Overall, Figure 2 indicates that Delta consistently identifies Guldager as the author of every segment extracted from Guldager2. However, for Guldager1, 16 of the 22 segments are incorrectly attributed to another writer. For example, the first segment is incorrectly attributed to Conradsen, while Guldager appears only in seventh position. Among these misattributions, Gazan frequently appears at the top of the ranking. As reported in Table 2, Guldager1 was previously associated with Gazan2. Interestingly, in the second part of Guldager1, six segments are correctly attributed to Guldager. This example illustrates that an author's stylistic characteristics may vary considerably throughout a long text.

## 4 Resampling on Delta

To improve the effectiveness of the Delta model, we propose applying a resampling procedure to the stylistic feature set. This approach is based on the bootstrap method (Efron & Tibshirani, 1994). As described previously, the stylistic feature set consists of the $m$ most frequent words (MFW) in the underlying corpus. The appropriate value of $m$, however, is not precisely defined and may vary depending on the researcher's choice.

In the proposed approach, different feature sets are generated by sampling with replacement from the original set of words. During this process, each term has an equal probability of being selected, namely $1/m$. Conversely, the probability that a given term is not selected in a single draw is $1 - (1/m)$. Since the sampling procedure is repeated $m$ times, the probability that a particular term is never selected is $[1 - (1/m)]^m$, which approaches approximately 0.368 (or $e^{-1}$) as $m$ increases. Consequently, each generated sample contains, on average, approximately 63.2% of the distinct terms in the original feature set. For example, when starting with $m = 316$ MFW, each resampled feature set will contain, on average, approximately 316 × 63.2% = 200 distinct terms.

When implementing this new strategy, one possibility would be to generate bootstrap samples in which stylistic features appearing more than once receive a weight proportional to their number of occurrences in the Z score calculation. However, we consider that this approach would assign excessive weight to certain stylistic features relative to others. As described in Equation 1, the Delta model already transforms the relative frequency of each word-type into a Z score to facilitate comparisons between stylistic profiles. From this perspective, multiplying the occurrence frequency before computing the Z score would distort the contribution of the feature to the representation of an author's stylistic idiosyncrasies.

In a second step, the Delta model is applied to each generated sample, and the resulting distance value is stored. After repeating this procedure $B$ times, we obtain $B$ distance values representing the variability associated with the computation of the stylistic distance between two documents. For this parameter, we suggest considering values ranging from 200 to 1,000.

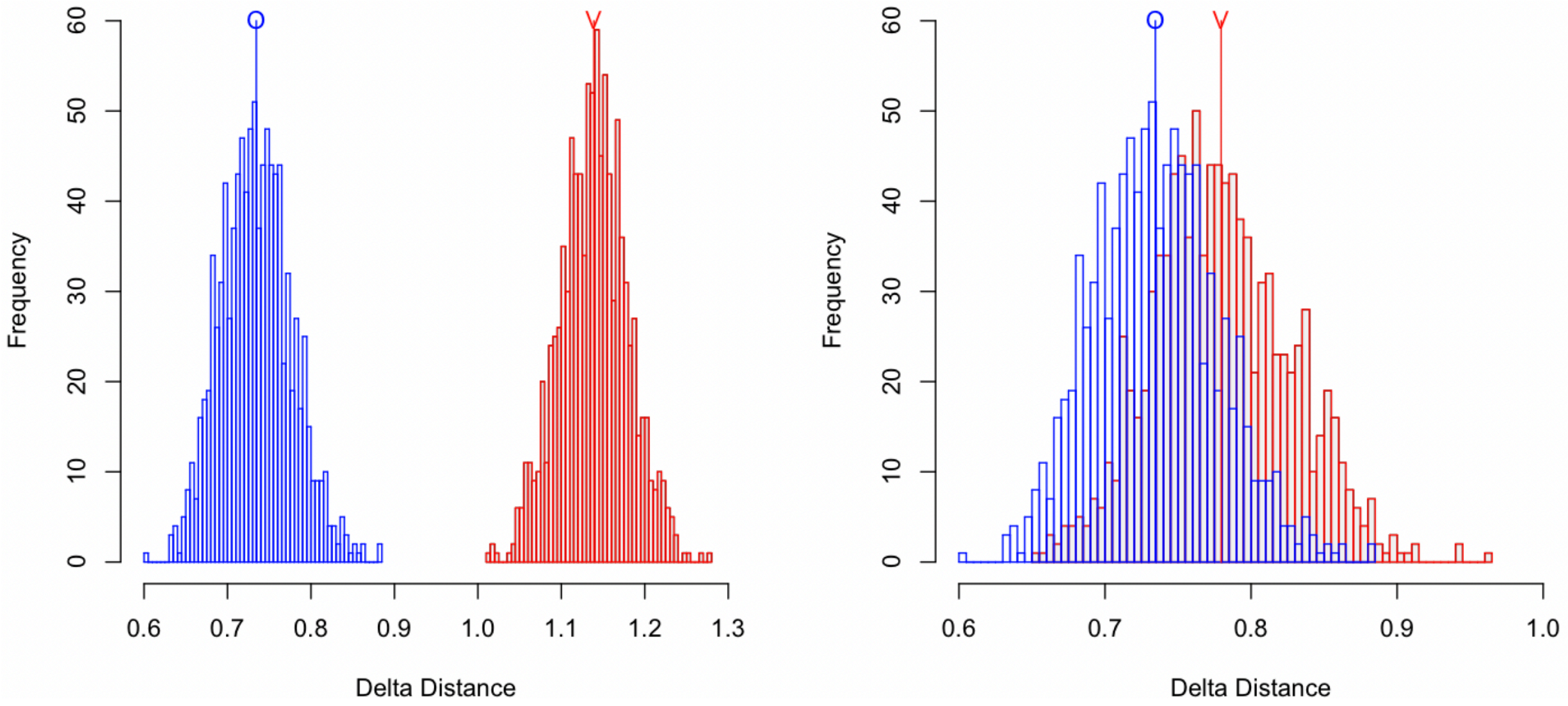

**Figure 3.** Histograms of Delta values using the resampling approach (316 MFW, B=1,000).
Left: Comparison of Guldager1-Guldager2 and Guldager1-Nordbo1.
Right: Comparison of Guldager1-Guldager2 and Guldager1-Gazan2.

Figure 3 displays three examples of histograms representing the resulting distance distributions. The histogram on the far left shows the distance values obtained when comparing the two novels written by Guldager. The mean distance is indicated by an ‘O’. Since the two documents were written by the same author, the resulting distance values are relatively small. The same graph also displays the distribution of distances between Guldager1 and Nordbo1, which clearly exhibits substantially larger values. The corresponding mean is indicated by a 'V'. These results suggest that Delta has no difficulty distinguishing between Guldager's and Nordbo's writing styles.

The graph on the right of Figure 3 displays the histograms of the distances between Guldager1 and Guldager2, and between Guldager1 and Gazan2. In this case, Delta has greater difficulty distinguishing between the two writing styles. However, when the mean values of the two distributions are considered (indicated by 'O' and 'V', respectively), the distinction between the two authors becomes clear. This figure illustrates the potential usefulness of our approach. Although the conclusion based on a particular feature set may be incorrect, considering the mean (or median) of the distance distribution can help identify each writer's stylistic idiosyncrasies.

Since *B* was set to 1,000 in this experiment, the difference between the mean and median is relatively small. For smaller values of *B*, we recommend using the median, which is generally considered a more robust statistic for assessing overall performance (Savoy, 1997).

**Table 3.** Author rankings obtained using Delta resampling (316 MFW, B = 500)

| \ Rank | 1 | | 2 | | 3 | |
|---|---|---|---|---|---|---|
| Novel | Novel | Distance | Novel | Distance | Novel | Distance |
| Blædel1 | Blædel2 | 0.554 | Gazan2 | 0.848 | Jessen2 | 0.863 |
| Blædel2 | Blædel1 | 0.559 | Gazan2 | 0.861 | Jessen2 | 0.907 |
| Conradsen1 | Conradsen2 | 0.579 | Guldager1 | 0.752 | Gazan | 0.885 |
| Conradsen2 | Conradsen1 | 0.585 | Guldager1 | 0.813 | Jessen2 | 0.892 |
| Gazan1 | Gazan2 | 0.703 | Jessen2 | 0.862 | Jessen1 | 0.892 |
| Gazan2 | Gazan1 | 1.328 | Halfdan1 | 1.395 | Guldager1 | 1.404 |
| Gerhardt1 | Gerhardt2 | 0.876 | Jessen1 | 0.982 | Tellerup2 | 0.997 |
| Gerhardt2 | Gerhardt1 | 0.897 | Jessen1 | 1.060 | Jessen2 | 1.066 |
| Guldager1 | Guldager2 | 0.738 | Conradsen1 | 0.762 | Gazan | 0.786 |
| Guldager2 | Guldager1 | 0.744 | Conradsen1 | 0.969 | Petersen1 | 1.013 |
| Halfdan1 | Halfdan2 | 0.425 | Gazan2 | 0.761 | Harvig1 | 0.802 |
| Halfdan2 | Halfdan1 | 0.429 | Sabroe1 | 0.783 | Harvig1 | 0.803 |
| Harvig1 | Harvig2 | 0.460 | Halfdan2 | 0.816 | Halfdan1 | 0.820 |
| Harvig2 | Harvig1 | 0.456 | Halfdan2 | 0.821 | Halfdan1 | 0.843 |
| Høeg1 | Høeg2 | 0.972 | Tellerup2 | 1.020 | Petersen1 | 1.030 |
| Høeg2 | Høeg1 | 1.028 | Petersen1 | 1.294 | Tellerup2 | 1.295 |
| Jessen1 | Jessen2 | 0.619 | Gazan2 | 0.812 | Korsgaard1 | 0.871 |
| Jessen2 | Jessen1 | 0.621 | Gazan2 | 0.809 | Gazan | 0.854 |
| Korsgaard1 | Korsgaard2 | 0.466 | Jessen1 | 0.886 | Gazan | 0.890 |
| Korsgaard2 | Korsgaard1 | 0.467 | Gazan2 | 0.925 | Jessen1 | 0.927 |
| Nordbo1 | Nordbo2 | 0.457 | Nordbo3 | 0.514 | Halfdan1 | 0.832 |
| Nordbo2 | Nordbo3 | 0.607 | Nordbo1 | 0.618 | Halfdan1 | 0.996 |
| Nordbo3 | Nordbo2 | 0.477 | Nordbo1 | 0.517 | Halfdan2 | 0.870 |
| Petersen1 | Petersen2 | 0.690 | Jessen1 | 0.883 | Halfdan1 | 0.955 |
| Petersen2 | Petersen1 | 0.693 | Tellerup2 | 1.023 | Tellerup2 | 1.023 |
| Sabroe2 | Sabroe1 | 0.421 | Halfdan2 | 0.842 | Halfdan1 | 0.906 |
| Tellerup1 | Tellerup2 | 0.798 | Høeg1 | 1.102 | Høeg1 | 1.102 |
| Tellerup2 | Tellerup1 | 0.780 | Høeg1 | 1.002 | Gerhardt1 | 1.011 |
| Blædel1 | Blædel2 | 0.554 | Gazan2 | 0.848 | Jessen2 | 0.863 |

To evaluate the proposed approach on the entire corpus, the 316 most frequent words (MFWs) were used to construct the original feature set. In this setting, each generated sample contains approximately 200 distinct words (316 × 63.2%), and the resulting distance values are similar to those reported in Table 2. The results of this experiment are presented in Table 3. The proposed method successfully eliminates the two incorrect assignments identified previously.

Similarly, to reproduce the experimental settings using 100, 250, 300, 500, and 800 MFWs, the number of MFWs was adjusted to 158, 395, 474, 790, and 1,264, respectively. Across these five additional experiments, the classical Delta method produces six incorrect assignments, whereas the resampling approach reduces this number to three. Clearly, the proposed technique does not eliminate all incorrect attributions produced by Delta; nevertheless, it appears useful for reducing the overall error rate.

**Table 4.** Ranking of the incorrect classification with the Rolling Delta

| Chunk # | 1 | 2 | 3 | 4 | 5 | 6 | 7 | 8 | 9 | 10 | 11 | 12 | 13 | 19 | 21 | 22 |
|---|---|---|---|---|---|---|---|---|---|---|---|---|---|---|---|---|
| Delta | 7 | 3 | 3 | 5 | 10 | 2 | 7 | 7 | 6 | 5 | 3 | 4 | 3 | 3 | 5 | 3 |
| Resampling | 5 | 3 | 2 | 5 | 7 | 2 | 4 | 3 | 5 | 3 | 2 | 3 | 3 | 2 | 2 | 2 |

The proposed resampling technique was also applied in conjunction with Rolling Delta. Table 4 reports the results obtained for the 22 passages extracted from Guldager1. For the 16 passages that were incorrectly attributed by the classical Delta method, the rank assigned to the second book by Guldager is reported in the row labeled ‘Delta’. For example, for the fourth segment, Delta ranked Guldager2 at the fifth position, whereas for the final segment, the correct answer appeared in third position.

Using the same segmentation of *Guldager1*, the resampling technique was then applied under identical conditions. The rank assigned to the correct book is reported in the row labeled “Resampling”. In this case, the proposed approach did not always rank the correct book first. However, as shown in Table 4, it generally placed the correct novel higher in the ranking than the classical method. To quantify the effect of the resampling technique, the mean rank of the correct book was 3.7 with Delta, compared with 2.7 with the proposed approach. Thus, an average improvement of one rank position was achieved. For the remaining passages, which are not reported in Table 4, both approaches ranked the correct book first.

## 5 Conclusion

When dealing with important authorship attribution cases, conclusions must be reached with a high degree of confidence and justified in plain English on the basis of stylistic characteristics. To achieve this, the Delta model offers several advantages. First, it has been applied in various contexts and has demonstrated a relatively low error rate. Second, an analysis of the Z scores makes it possible to explain the rationale behind each proposed attribution. Nevertheless, the estimated accuracy of Delta does not reach the level of certainty provided by DNA testing.

This study proposes a new technique for improving the effectiveness of Delta. More specifically, we propose resampling the feature set to generate $B$ Delta values (e.g., between 200 and 1,000). The median or mean of these values is then used as the final Delta distance between two documents.

Several evaluations of this approach indicate that the overall quality of the results can be improved compared with classical Delta. However, the resulting performance does not reach 100% accuracy, and some incorrect attributions remain. Finally, although the proposed modification has been developed in the context of the Delta model, it can also be applied to other authorship attribution methods.

## Disclosure Statement

No potential conflict of interest was reported by the author.

## Funding

No funding was received for this study.

## Appendix

**Table A.1.** Detailed information about the Danish corpus

| Author | Title | Subgenre | Publication year | Length (in tokens) |
|---|---|---|---|---|
| Blædel1 | *Bedemandens datter* | Suspense / Family | 2016 | 108,271 |
| Blædel2 | *Pigen under træet* | Crime | 2019 | 105,055 |
| Conradsen1 | *Centrifugen* | Crime | 2017 | 130,871 |
| Conradsen2 | *Familien Bruchfontaine* | Family saga | 2019 | 113,969 |
| Gazan1 | *Kvartsgeode* | Science / Suspense | 2020 | 16,958 |
| Gazan2 | *Svalens graf* | Science / Suspense | 2013 | 197,333 |
| Gerhardt1 | *Der bor Hollywoodstjerner på vejen* | Autofiction | 2014 | 42,593 |
| Gerhardt2 | *Transfervindue – fortællinger om de raskes fejl* | Autofiction | 2017 | 14,034 |
| Guldager1 | *En uskyldig familie* | Family | 2017 | 112,885 |
| Guldager2 | *Et rigtigt liv* | Family | 2019 | 81,597 |
| Halfdan1 | *Et nødvendigt offe* | Thriller | 2015 | 142,380 |
| Halfdan2 | *De uskyldige døde* | Crime | 2017 | 139,226 |
| Harvig1 | *Bedrag* | Crime / Psycho. | 2011 | 96,960 |
| Harvig2 | *Manden der vendte tilbage* | Crime / Psycho | 2017 | 89,610 |
| Høeg1 | *Tour de chambre* | Contemporary | 2020 | 27,176 |
| Høeg2 | *Nye rejsende* | Contemporary | 2017 | 13,603 |
| Jessen1 | *Doktor Bagges anagrammer* | Contemporary | 2017 | 35,229 |
| Jessen2 | *En ny tid* | Historical | 2015 | 51,787 |
| Korsgaard1 | *Hvis der skulle komme et menneske forbi* | Contemporary | 2017 | 85,141 |
| Korsgaard2 | *En dag vil vi grine af det* | Contemporary | 2018 | 80,493 |
| Nordbo1 | *Pigen uden hud* | Crime / Thriller | 2017 | 110,748 |
| Nordbo2 | *Kold angst* | Crime / Thriller | 2018 | 112,535 |
| Nordbo3 | *Kvinden med dødsmasken* | Crime / Thriller | 2019 | 101,847 |
| Petersen1 | *Værelse til leje* | Crime | 2017 | 87,761 |
| Petersen2 | *Ønskebarnet* | Psycho. Suspense | 2009 | 95,913 |
| Sabroe1 | *32 centimeter* | Contemporary | 2018 | 114,540 |
| Sabroe2 | *Love Me Do* | Contemporary | 2017 | 107,886 |
| Tellerup1 | *Markløs* | Contemporary | 2017 | 47,788 |
| Tellerup2 | *Hedeselskabet* | Contemporary | 2020 | 51,291 |